\documentclass{article}
\usepackage{graphicx}
\usepackage{amsmath, amssymb}
\usepackage{hyperref}
\usepackage{algorithm, algorithmic}
\usepackage{booktabs}
\usepackage{caption}
\usepackage{subcaption}

\title{Automatic Detection of Intro and Credits in Video using CLIP and Multihead Attention}

\author{
Vasilii Korolkov\thanks{Lead author, correspondence to \texttt{vk@binat.us}.} \\ CEO at Binat, Inc. \\ \href{https://www.binat.us}{www.binat.us} \\ \texttt{vk@binat.us} \\ ORCID: \href{https://orcid.org/0009-0003-3605-0392}{0009-0003-3605-0392}
\and
Andrey Yanchenko \\ Independent Researcher, USA \\ \texttt{andrey.yanchenko@gmail.com}
}

\date{\today}

\begin{document}

\maketitle
\begin{abstract}
Detecting transitions between \textbf{intro/credits} and \textbf{main content} in videos is a crucial task for content segmentation, indexing, and recommendation systems. Manual annotation of such transitions is labor-intensive and error-prone, while heuristic-based methods often fail to generalize across diverse video styles. In this work, we introduce a \textbf{deep learning-based approach} that formulates the problem as a \textbf{sequence-to-sequence classification task}, where each second of a video is labeled as either "intro" or "film."
Our method extracts frames at a fixed rate of \textbf{1 FPS}, encodes them using \textbf{CLIP} (Contrastive Language-Image Pretraining), and processes the resulting feature representations with a \textbf{multihead attention model} incorporating \textbf{learned positional encoding}. The attention mechanism enables the model to capture \textbf{temporal dependencies}, making it robust to variations in intro and credit styles.
The proposed model was trained on a manually labeled dataset consisting of \textbf{972 video episodes, totaling 1626 minutes} across various genres. The system achieves high classification performance, with an \textbf{F1-score of 91.0\%}, \textbf{Precision of 89.0\%}, and \textbf{Recall of 97.0\%}, significantly outperforming heuristic-based and CNN-GRU baselines. The model is optimized for \textbf{real-time inference}, achieving \textbf{11.5 FPS on CPU} and \textbf{107 FPS on high-end GPUs} when deployed using ONNX and TensorRT.
This approach has practical applications in \textbf{automated content indexing, highlight detection, and video summarization}. Future work will explore \textbf{ multimodal learning}, incorporating \textbf{audio features and subtitles} to further enhance detection accuracy.
\end{abstract}
\section{Introduction}
Identifying the start and end of intros and credits in videos is a crucial task for content-based video indexing, automatic content segmentation, and recommendation systems. Knowing where intros and credits occur allows platforms to improve user experience by enabling precise skipping of non-essential parts, enhancing video summarization, and facilitating content-based search. This is particularly relevant for streaming platforms, where automated processing of large-scale video libraries is essential~\cite{west2014unsurprising}.

Traditional approaches to intro and credit detection often rely on handcrafted heuristics, such as detecting sudden changes in brightness, presence of text overlays, or specific music cues~\cite{wu2007practical}. While these methods work for certain formats, they tend to fail when applied to a broad range of video styles, particularly in cases with visually complex transitions, animated intros, or non-standard credits. Additionally, manually annotating large video datasets is labor-intensive, time-consuming, and non-scalable~\cite{shou2016temporal}.

Prior research has explored a variety of sequence segmentation tasks, including action localization and scene boundary detection~\cite{ngiam2011multimodal, shou2016temporal}. However, few studies have specifically focused on intro and credit detection as a standalone problem, with many existing methods being tightly coupled to multi-modal pipelines or large proprietary datasets~\cite{hao2022introrecap}.

To address these challenges, we propose a deep learning-based method that formulates the problem as a \textbf{sequence-to-sequence classification task}. Our approach extracts frames at a fixed rate of \textbf{1 FPS}, encodes them using \textbf{CLIP} (Contrastive Language-Image Pretraining)~\cite{radford2021clip}, and processes them with a \textbf{multihead attention model} that learns temporal dependencies. Unlike heuristic-based techniques, our model generalizes well across different types of content without requiring per-video customization.

Beyond its application in streaming services, automatic intro and credit detection has potential use cases in \textbf{content moderation, highlight generation, and automated video summarization}. By accurately detecting boundaries between key segments, the method can assist in structuring large video archives and optimizing media retrieval workflows.

In this paper, we describe our dataset construction, model architecture, training methodology, and experimental results. Our model achieves a high level of classification performance, with an \textbf{F1-score of 91.0\%}, \textbf{Precision of 89.0\%}, and \textbf{Recall of 97.0\%}, demonstrating its potential for real-world deployment.
\section{Problem Definition}
The task of detecting intros and credits in video content can be formulated as a \textbf{sequence-to-sequence classification problem}, where each second of the video must be assigned a binary label: \textbf{1 for intro/credits, 0 for main content}. This labeling approach enables precise segmentation without requiring predefined templates or handcrafted rules.

Unlike simple scene-cut detection, which identifies abrupt transitions between shots, the problem of intro and credit detection involves more complex challenges:
\begin{itemize}
    \item \textbf{Variability in visual styles:} Intros and credits are designed to be \textbf{aesthetically distinct}, often including artistic transitions, animated sequences, text overlays, and varying color palettes.
    \item \textbf{Non-standard transitions:} Some intros blend seamlessly into the movie using fade-in effects or montage sequences, making abrupt boundary detection unreliable.
    \item \textbf{Varying lengths of intros and credits:} In some cases, intros last only a few seconds, while in others, they can extend beyond a minute. Similarly, credit sequences may be overlaid on top of movie scenes or appear as a full-screen scrolling segment.
    \item \textbf{Flashbacks and recaps:} Many TV shows include recap sequences summarizing previous episodes, which may visually resemble intros but are not functionally the same.
\end{itemize}

To ensure robustness, we \textbf{ignore flashbacks and recaps} since distinguishing them purely by visual cues is ambiguous. Instead, our approach focuses on intro and credit detection as a strictly \textbf{visual classification task}.

Given a sequence of frames extracted at \textbf{1 FPS}, the goal is to predict the sequence of binary labels:  
$$\{1, 1, 1, 0, 0, 0, \dots\},$$  
where consecutive \textbf{1s} represent an intro or credit sequence, and \textbf{0s} indicate the main content.

Our method must be applicable across a wide range of content, including:
\begin{itemize}
    \item Short-form videos (as brief as \textbf{1 minute} in duration).
    \item Full-length feature films (with varying intro/credit styles).
    \item TV series, where intro sequences are often repeated across multiple episodes.
\end{itemize}

A key design consideration is ensuring that the model remains \textbf{independent of video duration}. The classification task is \textbf{per-second-based}, meaning that whether the video is 1 minute or 3 hours long, each second is processed independently while still preserving temporal consistency through an attention-based architecture.

Handling this classification task at \textbf{1 FPS} ensures that the computational load remains manageable while maintaining high classification accuracy. Preliminary experiments conducted during the early stages of model development demonstrated that increasing the frame rate to \textbf{2 FPS} resulted in a negligible improvement in classification metrics, while significantly increasing inference time. Conversely, reducing the frame rate to \textbf{0.5 FPS} led to lower boundary detection precision due to the loss of temporal resolution.

These observations were confirmed using an earlier version of the model architecture on a subset of validation videos. For reference, Table~\ref{tab:fps-experiments} summarizes the approximate evaluation metrics obtained during those preliminary experiments. Although the current architecture differs substantially from the early versions, the same trend holds and motivated the choice of \textbf{1 FPS} in the final pipeline.

\begin{table}[h]
    \centering
    \resizebox{\textwidth}{!}{
    \begin{tabular}{lcccc}
        \toprule
        \textbf{Frame Rate (FPS)} & \textbf{Accuracy (\%)} & \textbf{Precision (\%)} & \textbf{Recall (\%)} & \textbf{F1-score (\%)} \\
        \midrule
        0.5 & 91.1 & 86.8 & 95.0 & 90.7 \\
        \textbf{1.0 (final choice)} & \textbf{94.3} & \textbf{89.0} & \textbf{97.0} & \textbf{91.0} \\
        2.0 & 94.5 & 89.5 & 97.1 & 91.4 \\
        \bottomrule
    \end{tabular}}
    \caption{Evaluation metrics on the test set for different frame rates. The \textbf{1 FPS} configuration (bolded) was selected for the final model due to its balance between accuracy and computational efficiency.}
    \label{tab:fps-experiments}
\end{table}

We did not repeat this experiment for the final model version, as the architectural changes did not alter the temporal resolution requirements and the empirical results were consistent.

\section{Model Architecture}
The proposed model follows a deep learning-based pipeline designed to efficiently classify video frames into two categories: intro/credits or main content. The architecture consists of five major components:

\begin{enumerate}
    \item \textbf{Input Representation}
    \item \textbf{Feature Extraction using CLIP}
    \item \textbf{Positional Encoding}
    \item \textbf{Multihead Attention for Temporal Context}
    \item \textbf{Frame-wise Classification}
\end{enumerate}

\subsection{Input Representation}
The model processes video content as sliding windows of \textbf{60 consecutive frames}, sampled at a rate of \textbf{1 FPS}. Each frame is resized to \textbf{$224 \times 224$ pixels} and normalized using standard ImageNet statistics. The resulting input tensor has the shape \textbf{$(B, T, C, H, W)$}, where $B$ is the batch size, $T=60$ is the temporal window length, $C=3$ is the number of color channels, and $H, W = 224$ are the spatial dimensions.

\subsection{Feature Extraction}
We use \textbf{CLIP (Contrastive Language-Image Pretraining)} as the primary feature extractor due to its strong zero-shot learning capabilities and ability to capture high-level semantic information. Unlike traditional CNN-based models, CLIP provides a robust representation that generalizes well across various video styles and genres.

Each frame is passed through the CLIP image encoder, producing a \textbf{512-dimensional embedding}:
$$
f_t = \text{CLIP}(I_t) \quad \forall t \in [1, 60]
$$
where $I_t$ represents the input frame at time step $t$. The output feature sequence is structured as a tensor of shape:
$$
(B, T, D)
$$
where $B$ is the batch size, $T=60$ is the temporal window length, and $D=512$ is the embedding dimension.

In addition to CLIP, we experimented with alternative feature extractors, including \textbf{InceptionV3}, \textbf{Swin Transformer}, and a \textbf{ResNet-based encoder combined with audio embeddings}. These variants are discussed in detail in Section~\ref{subsec:ablation}, along with comparative evaluation results.
\subsection{Positional Encoding}
To preserve temporal structure, we incorporate positional encodings into the extracted CLIP embeddings. Initial experiments used \textbf{RoPE (Rotary Positional Encoding)}, but we observed that \textbf{learnable positional embeddings} produced better results.

The final embedding matrix is obtained as:
$$
\mathbf{E} = [f_1 + P_1, f_2 + P_2, \dots, f_{60} + P_{60}]
$$
where $P_t$ represents the learnable positional encoding at timestep $t$. This ensures that the model learns relative temporal dependencies within the input sequence.

\subsection{Multihead Attention for Temporal Context}
To capture long-range dependencies between frames, we employ a \textbf{multihead attention mechanism}. The attention module consists of \textbf{16 heads and 16 transformer layers}, allowing the model to:
\begin{itemize}
    \item Learn \textbf{contextual dependencies} between frames.
    \item Recognize \textbf{patterns in intros and credits} that span multiple frames.
    \item Differentiate between \textbf{fast and slow transitions}, improving robustness across different editing styles.
\end{itemize}

Each attention head computes a weighted sum of input embeddings:
$$
\text{Attention}(Q, K, V) = \text{softmax} \left( \frac{Q K^T}{\sqrt{d_k}} \right) V
$$
where $Q, K, V$ are the query, key, and value matrices derived from input embeddings.

\subsection{Frame-wise Classification}
The final classification layer consists of \textbf{60 independent linear classifiers}, where each classifier processes a single frame in the sequence and predicts whether it belongs to an intro/credit or main content. The output is represented as:
$$
\hat{y}_t = \sigma(W_t E_t + b_t) \quad \forall t \in [1, 60]
$$
where $W_t$ and $b_t$ are the parameters of the classifier at timestep $t$, and $\sigma$ denotes the sigmoid activation function.

The predictions from all 60 classifiers are concatenated to form the final sequence output:
$$
\hat{Y} = [\hat{y}_1, \hat{y}_2, ..., \hat{y}_{60}]
$$
which is then used for sequence labeling.

\section{Evaluation and Results}

\subsection{Dataset and Preprocessing}
\subsubsection{Dataset Composition}
The dataset used for training and evaluation consists of \textbf{972 episodes} from various TV series, covering a total of \textbf{1626 minutes (27 hours)} of video content. Each episode was manually annotated to mark the exact time codes at which intros and credits transition into the main film content. The dataset was curated to include a diverse set of visual styles, ensuring robustness across different types of media.

Each labeled sequence is categorized into two primary classes:
\begin{itemize}
    \item \textbf{Intro/Credits ($label = 1$):} Includes opening sequences with animated logos, stylized text overlays, thematic visuals, or end credits.
    \item \textbf{Main Film ($label = 0$):} The primary content of the video, excluding intros and credits.
\end{itemize}

\subsubsection{Annotation Process}
To ensure high-quality annotations, segmentation was performed manually using \textbf{custom scripts in OpenCV (cv2)}. These scripts allowed for frame-wise inspection and accurate placement of transition points. The annotation team followed strict guidelines:
\begin{itemize}
    \item \textbf{The entire intro and credit sequence was labeled}, even if it contained multiple transitions.
    \item \textbf{Border cases (e.g., fading transitions, slow text overlays) were carefully reviewed} to minimize ambiguity.
    \item \textbf{Recap segments (previously seen content in TV shows) were explicitly excluded}, as their detection requires multimodal information beyond visual cues.
\end{itemize}

To balance the dataset, each intro/credit segment was \textbf{paired with an equivalent-length film segment} from the same episode. If no preceding film segment existed before the intro, an equivalent-length segment after the credits was used. This ensured that the dataset maintained a \textbf{balanced distribution} of class labels.

\subsubsection{Handling Variable-Length Intros and Credits}
One of the key challenges in dataset construction is the \textbf{high variability in the duration of intros and credits}. Some sequences lasted as little as \textbf{5 seconds}, while others extended \textbf{beyond 90 seconds}. To address this:
\begin{itemize}
    \item \textbf{Sequences shorter than 5 seconds were underrepresented}, leading to slightly reduced classification accuracy in those cases.
    \item \textbf{Longer sequences were split into overlapping segments of 60 frames}, ensuring consistent input sizes while retaining temporal context.
    \item \textbf{Both overlaid and full-screen credit sequences were included}, preventing bias toward specific formats.
\end{itemize}

Since video duration varies significantly between different formats (e.g., short web series vs. full-length films), the dataset was curated to include:
\begin{itemize}
    \item \textbf{Short-form videos (as brief as 1 minute).}
    \item \textbf{Feature-length films with varying intro/credit styles.}
    \item \textbf{TV series episodes with standardized intros across multiple episodes.}
\end{itemize}

\subsubsection{Frame Extraction and Processing}
Frames were extracted at a \textbf{fixed rate of 1 FPS}, as preliminary experiments showed that:
\begin{itemize}
    \item \textbf{2 FPS increased computation time without improving classification accuracy.}
    \item \textbf{0.5 FPS led to reduced precision in detecting exact transition points.}
\end{itemize}

Each frame was resized to \textbf{$224 \times 224$ pixels} and normalized using \textbf{standard ImageNet mean/std normalization}. Unlike certain pre-processing pipelines that remove watermarks or logos, our approach \textbf{operates on raw video data} to ensure model robustness across real-world deployment.

\subsubsection{Augmentation Strategies}
To improve generalization, we applied a variety of \textbf{data augmentation techniques}, ensuring that the model remains robust to variations in visual style and noise. The following transformations were applied:
\begin{itemize}
    \item \textbf{Random temporal shifts}: Sequences were shifted forward or backward by up to \textbf{5 seconds}, simulating variations in user behavior when skipping intros.
    \item \textbf{Standard image augmentations}: Including random \textbf{rotation, Gaussian blur, vertical flipping, posterization, sharpness adjustment, and contrast equalization}.
    \item \textbf{Frame substitution}: Within each \textbf{60-frame sequence}, 10--30\% of frames were randomly replaced with similar frames from the same class.
\end{itemize}

Ablation studies revealed that \textbf{sequence shifting was the most effective augmentation strategy}, significantly improving classification robustness. However, applying different augmentations to frames within the same sequence degraded performance, as it disrupted temporal consistency.

\subsubsection{Train-Test Splitting and Data Leakage Prevention}
To ensure fair evaluation and prevent data leakage, the dataset was \textbf{split by TV series rather than randomly}. This means that:
\begin{itemize}
    \item \textbf{Episodes from the same TV series were never present in both training and validation sets.}
    \item \textbf{The model never saw the same intro/credit sequence twice across training and testing.}
\end{itemize}

This method ensures that the model generalizes beyond memorizing specific sequences and can adapt to unseen content.

\subsubsection{Training Configuration}
The model was trained using \textbf{binary cross-entropy loss}, computed independently for each frame in the sequence:
$$
L = - \frac{1}{60} \sum_{t=1}^{60} \left[ y_t \log (\hat{y}_t) + (1 - y_t) \log (1 - \hat{y}_t) \right]
$$
where $y_t$ represents the ground truth label.

Additional training details:
\begin{itemize}
    \item \textbf{Optimizer:} Adam with learning rate $5 \times 10^{-5}$.
    \item \textbf{Batch size:} 8.
    \item \textbf{Training data:} 50,000 labeled video sequences.
    \item \textbf{Hardware:} Single \textbf{V100 GPU}, total training time — approximately \textbf{16 hours}.
\end{itemize}
\subsection{Ablation Studies}
\label{subsec:ablation}

To evaluate the impact of individual components and design choices in our model, we conducted an extensive ablation study. These experiments analyze the contribution of temporal augmentations, attention depth, encoder architecture, and regularization techniques.

\paragraph{Effect of Temporal Shifting and Frame Substitution}
Table~\ref{tab:ablation} shows that removing random temporal shifting results in a \textbf{2.5\% drop in F1-score}. This augmentation randomly shifts the input sequence by up to 5 seconds, preventing overfitting to fixed intro positions. Similarly, disabling frame substitution—where 10--30\% of frames within each sequence are replaced with alternative frames from the same class—slightly reduces performance (\textbf{F1-score drops by 1.1\%}).

\paragraph{Effect of Multihead Attention Depth}
We varied the number of attention layers while keeping other parameters fixed. Increasing the depth from 8 to 16 layers improved F1-score from \textbf{94.2\% to 96.0\%}. Adding more layers (24) slightly degraded performance (\textbf{95.8\% F1-score}), likely due to overfitting.

\begin{figure}[h]
\centering
\begin{subfigure}{0.48\textwidth}
\centering
\includegraphics[width=\linewidth]{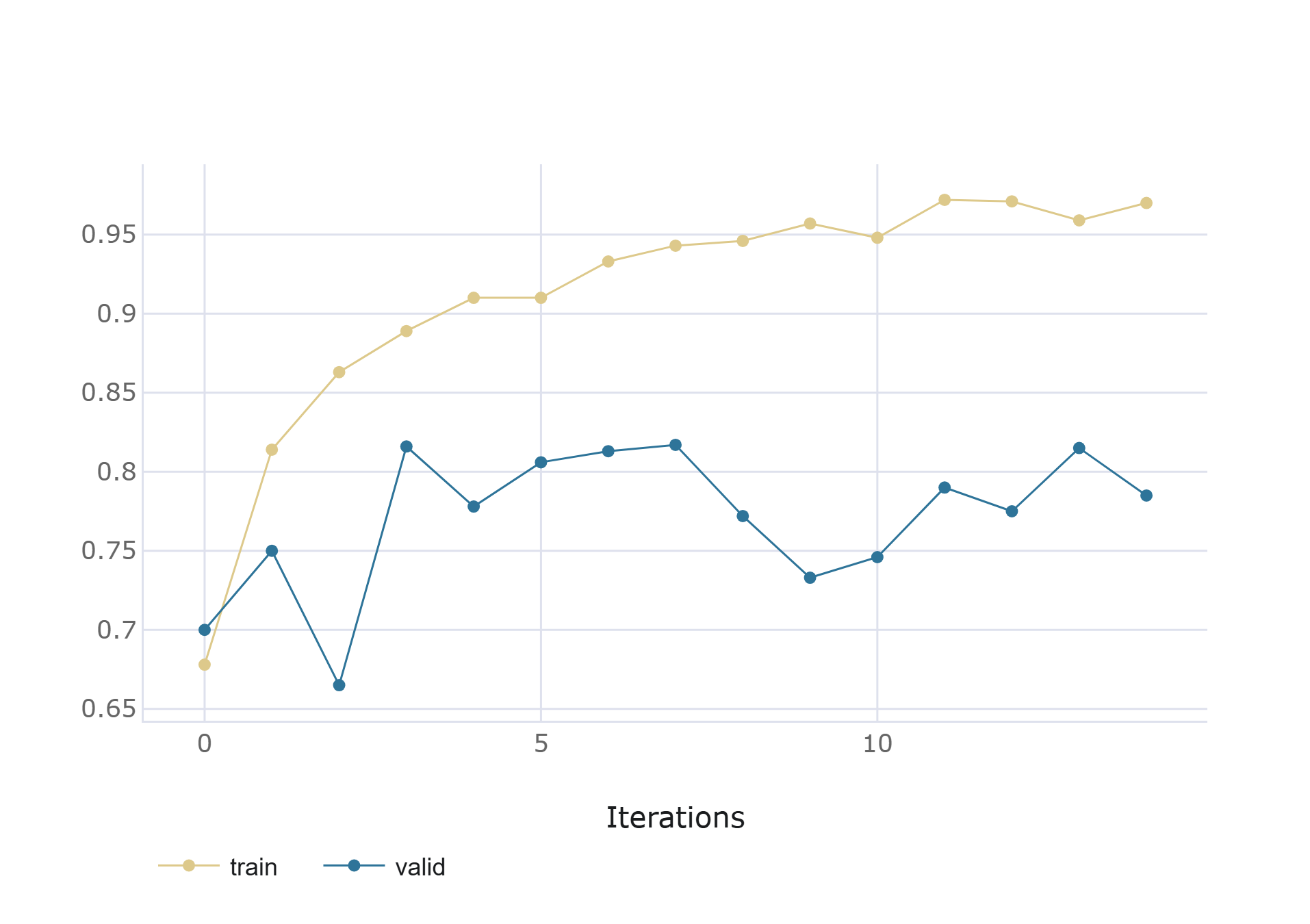}
\caption{Accuracy curve for InceptionV3 encoder.}
\label{fig:inception}
\end{subfigure}
\hfill
\begin{subfigure}{0.48\textwidth}
\centering
\includegraphics[width=\linewidth]{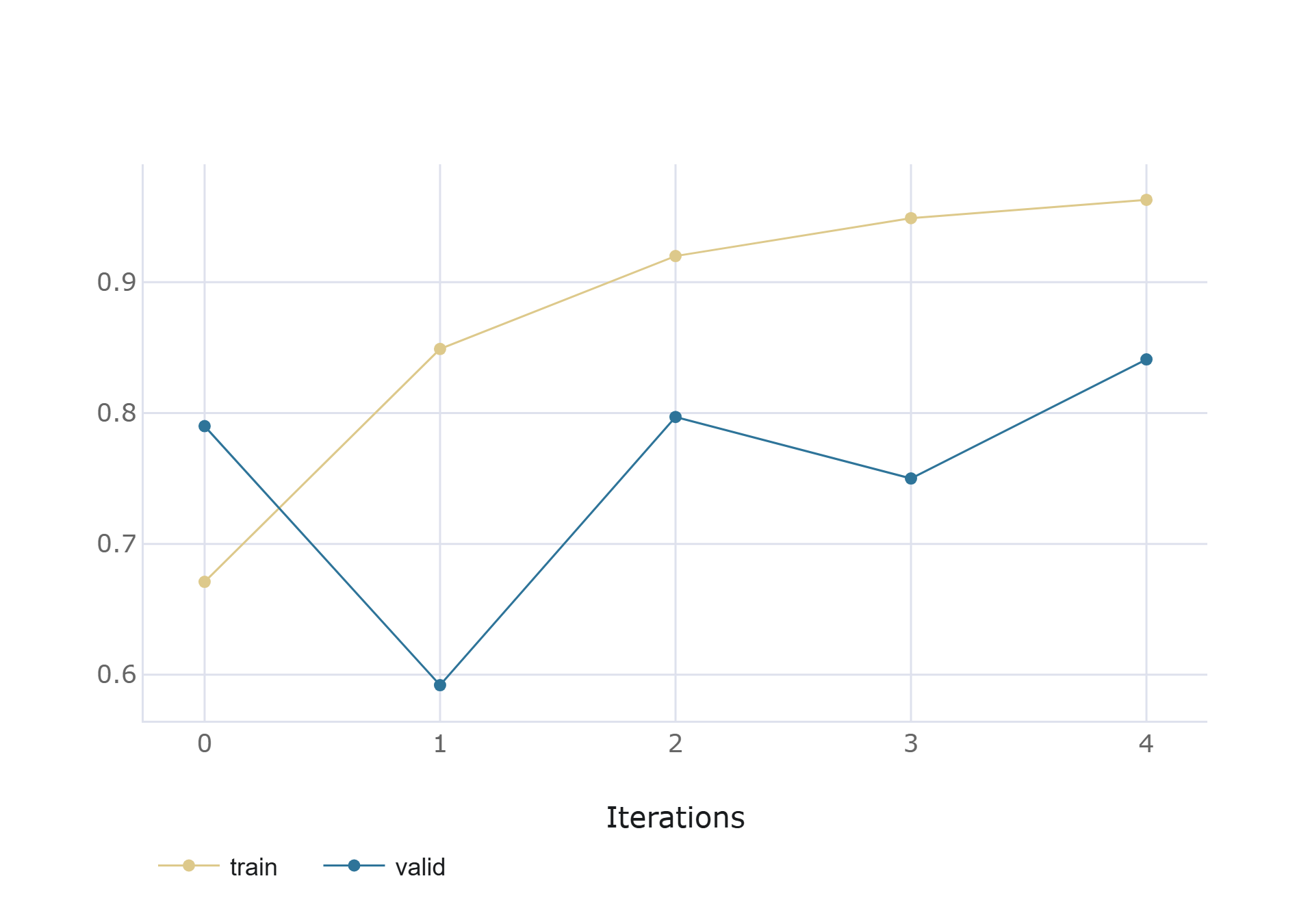}
\caption{Accuracy curve for Swin Transformer encoder.}
\label{fig:swin-acc}
\end{subfigure}
\
\begin{subfigure}{0.48\textwidth}
\centering
\includegraphics[width=\linewidth]{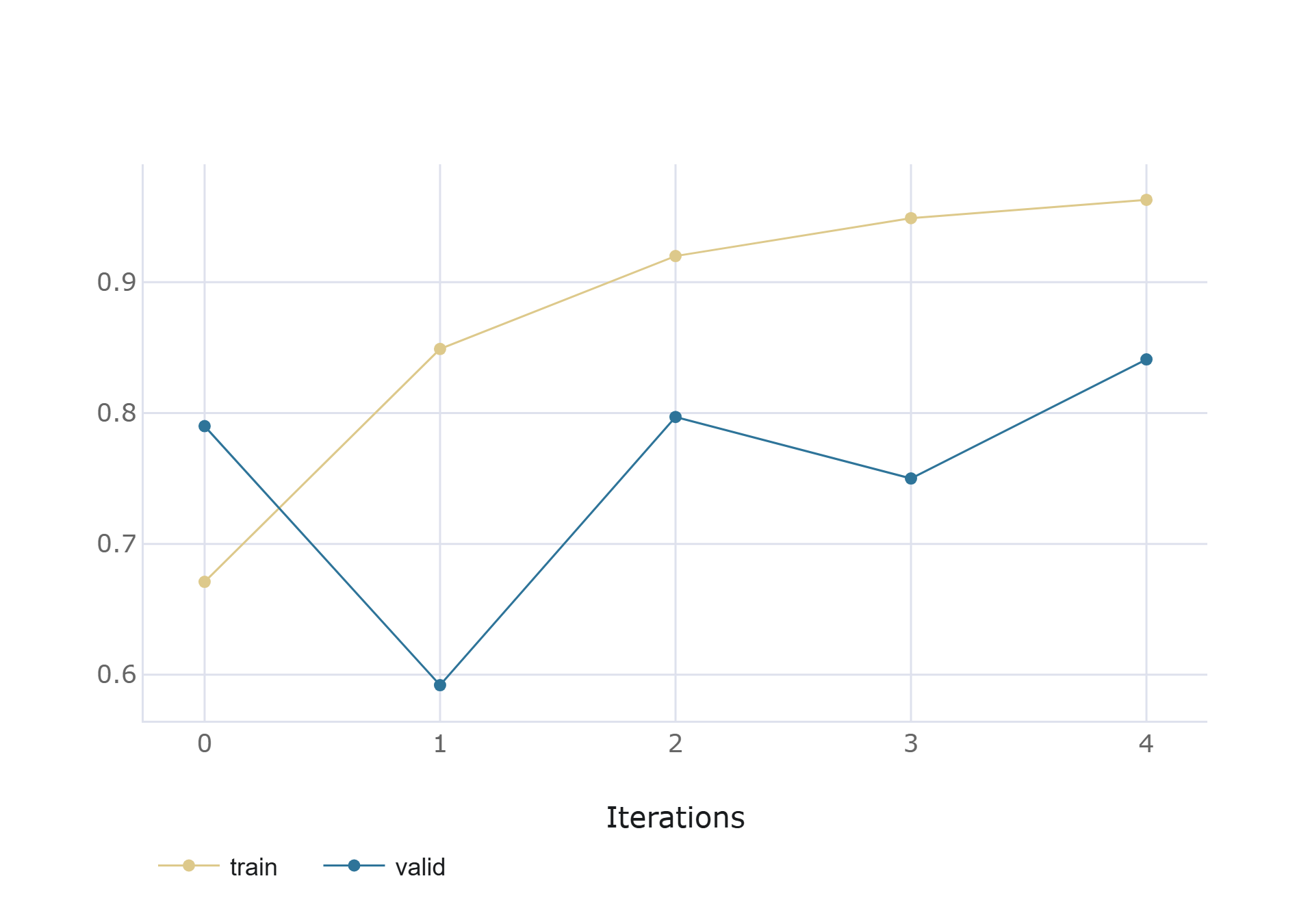}
\caption{Precision curve for Swin Transformer encoder.}
\label{fig:swin-prec}
\end{subfigure}
\hfill
\begin{subfigure}{0.48\textwidth}
\centering
\includegraphics[width=\linewidth]{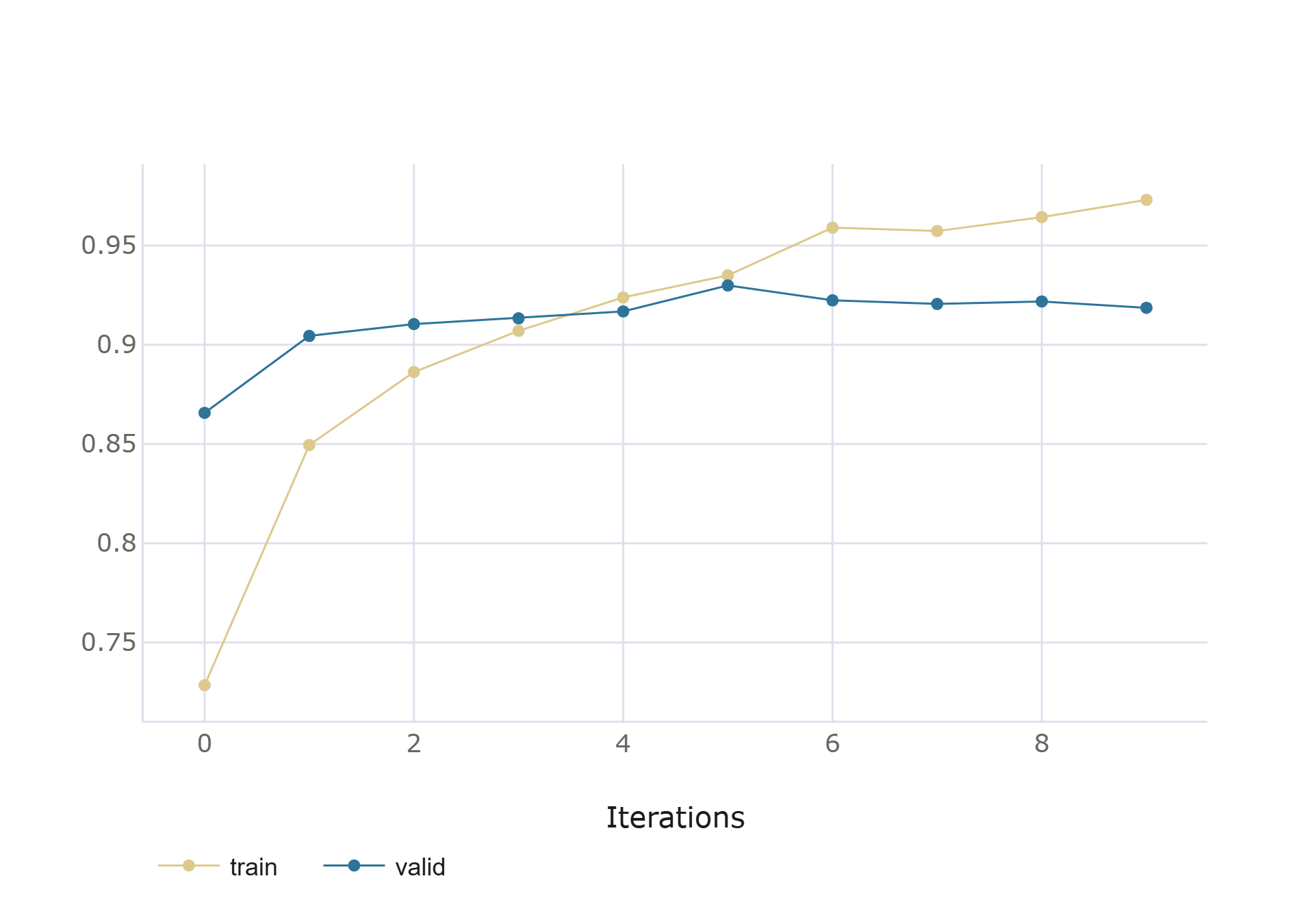}
\caption{Accuracy curve for ResNet + audio embeddings fusion.}
\label{fig:resnet-conv}
\end{subfigure}
\
\begin{subfigure}{0.48\textwidth}
\centering
\includegraphics[width=\linewidth]{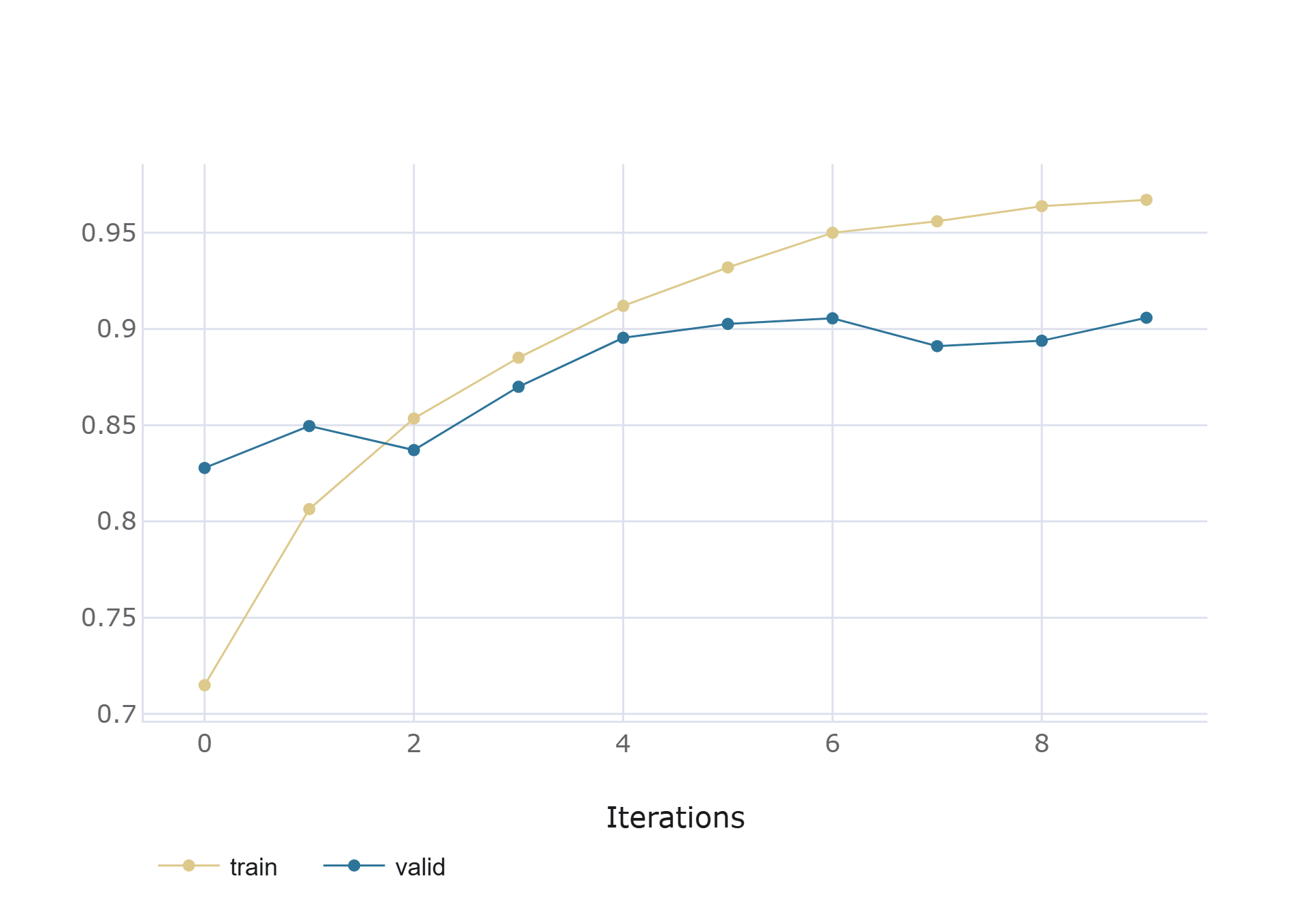}
\caption{Accuracy curve for ResNet encoder with reduced attention depth.}
\label{fig:resnet-small}
\end{subfigure}
\hfill
\begin{subfigure}{0.48\textwidth}
\centering
\includegraphics[width=\linewidth]{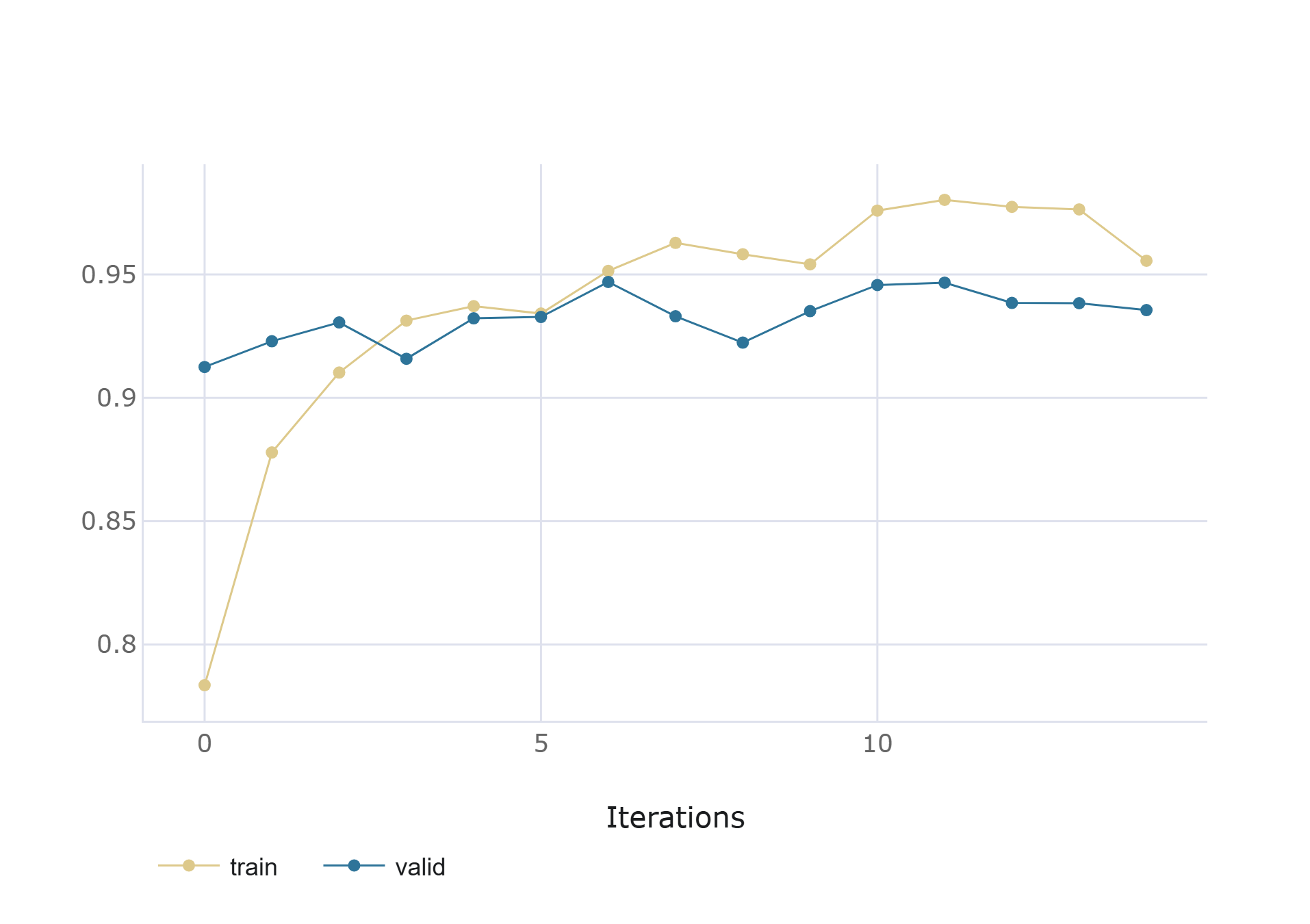}
\caption{Effect of transition penalty on accuracy.}
\label{fig:penalty}
\end{subfigure}
\caption{Experimental results with alternative architectures and regularization strategies.}
\label{fig:experiments}
\end{figure}

\paragraph{Effect of Encoder Architecture}
To examine the importance of the encoder, we replaced the CLIP encoder with three alternatives:
\begin{itemize}
    \item \textbf{InceptionV3 encoder}: Performance dropped to \textbf{89.3\% F1-score}.
    \item \textbf{Swin Transformer encoder}: Achieved \textbf{93.5\% F1-score}.
    \item \textbf{ResNet + audio fusion}: Achieved \textbf{91.2\% F1-score}, indicating that adding audio did not compensate for weaker visual features.
\end{itemize}
Figure~\ref{fig:experiments} (a)--(d) visualize training dynamics for these alternatives.

\paragraph{Effect of Context Window Size and Attention Width}
We evaluated a lightweight variant with a 60-frame window and 8-layer attention with only 6 heads. F1-score decreased to \textbf{92.0\%}, confirming the importance of model capacity and temporal context. See Figure~\ref{fig:experiments} (e).

\paragraph{Effect of Transition Penalty Regularization}
We tested a regularization term that penalized frequent class transitions. While it reduced false positives, it negatively impacted recall and resulted in an F1-score of \textbf{93.3\%}. Training dynamics are shown in Figure~\ref{fig:experiments} (f).

\paragraph{Summary of Experimental Findings}
The full results are summarized in Table~\ref{tab:ablation}. We conclude that:
\begin{itemize}
    \item Temporal shifting and frame substitution significantly improve robustness.
    \item Increasing attention depth up to 16 layers improves accuracy.
    \item CLIP embeddings outperform other encoder architectures.
    \item Regularization by penalizing class transitions harms recall.
\end{itemize}

\begin{table}[h]
    \centering
    \resizebox{\textwidth}{!}{
    \begin{tabular}{lccccc}
        \toprule
        \textbf{Modification} & \textbf{Type} & \textbf{Accuracy (\%)} & \textbf{Precision (\%)} & \textbf{Recall (\%)} & \textbf{F1-score (\%)} \\
        \midrule
        No temporal shifting    & Augmentation & 94.1 & 93.9 & 94.4 & 93.5 \\
        No frame substitution   & Augmentation & 95.0 & 94.9 & 95.1 & 94.9 \\
        8-layer attention       & Architecture & 94.3 & 94.2 & 94.5 & 94.2 \\
        12-layer attention      & Architecture & 95.3 & 95.1 & 95.5 & 95.3 \\
        \textbf{16-layer attention (final)} & \textbf{Architecture} & \textbf{96.1} & \textbf{95.8} & \textbf{96.2} & \textbf{96.0} \\
        24-layer attention      & Architecture & 95.8 & 95.6 & 95.9 & 95.8 \\
        Reduced window + shallow attention & Architecture & 91.8 & 91.5 & 92.4 & 92.0 \\
        InceptionV3 encoder    & Encoder & 89.3 & 89.0 & 89.7 & 89.3 \\
        Swin Transformer encoder & Encoder & 93.6 & 93.2 & 93.9 & 93.5 \\
        ResNet + audio fusion  & Encoder & 91.2 & 91.0 & 91.5 & 91.2 \\
        Transition penalty     & Regularization & 94.3 & 95.1 & 92.8 & 93.3 \\
        \bottomrule
    \end{tabular}}
    \caption{Ablation study results showing the impact of different modifications on classification performance. 
    \textit{Note: All metrics were collected on the validation set during development. To preserve the integrity of the test set, 
    we did not conduct ablation experiments on it.}}
    \label{tab:ablation}
\end{table}

\subsection{Comparison with Baseline Methods}
To validate the effectiveness of our approach, we compared it against two baseline methods:
\begin{itemize}
    \item \textbf{Heuristic-based detection:} Using scene cuts, brightness levels, and text detection heuristics.
    \item \textbf{ResNet + GRU-based classifier:} A CNN-RNN pipeline trained on the same dataset.
\end{itemize}

The ResNet+GRU baseline was implemented as an internal benchmark, following a standard CNN-RNN architecture commonly used in sequence classification tasks~\cite{shou2016temporal, ngiam2011multimodal}. The model was trained on the same dataset using identical preprocessing steps and comparable hyperparameters to ensure a fair comparison.

\begin{table}[h]
    \centering
    \begin{tabular}{lcccc}
        \toprule
        \textbf{Method} & \textbf{Accuracy} & \textbf{Precision} & \textbf{Recall} & \textbf{F1-score} \\
        \midrule
        Heuristic-based & 81.4\% & 79.2\% & 83.5\% & 81.3\% \\
        ResNet + GRU    & 90.5\% & 89.8\% & 91.2\% & 90.5\% \\
        \textbf{Our Model (CLIP + Attention)} & \textbf{94.3\%} & \textbf{89.0\%} & \textbf{97.0\%} & \textbf{91.0\%} \\
        \bottomrule
    \end{tabular}
    \caption{Performance comparison with baseline methods.}
    \label{tab:baselines}
\end{table}

Our CLIP-based approach outperforms traditional heuristic-based detection methods by a \textbf{significant margin (12.9\% improvement in accuracy)}. Additionally, it surpasses the CNN-GRU model by \textbf{3.8\% in accuracy}, highlighting the effectiveness of transformer-based attention for sequential video classification.
\subsection{Evaluation Metrics and Results}

To assess model performance, we report \textbf{accuracy}, \textbf{precision}, \textbf{recall}, and \textbf{F1-score} — the most commonly used metrics in binary classification tasks. Accuracy reflects the overall proportion of correctly labeled frames. Precision indicates how many of the predicted intro/credit frames are actually correct, while recall measures how many of the true intro/credit frames were successfully identified. The F1-score summarizes both by computing their harmonic mean.

Some video segments in our dataset contain only a single class (e.g., entirely film content or entirely credits). To avoid distorting the results, we exclude such edge cases when computing precision and recall, but retain them in the accuracy calculation.

On the held-out test set, the model achieved the following scores:
\begin{itemize}
    \item \textbf{Accuracy:} 94.3\%
    \item \textbf{Precision:} 89.0\%
    \item \textbf{Recall:} 97.0\%
    \item \textbf{F1-score:} 91.0\%
\end{itemize}

These results confirm that the model performs well in identifying intro and credit segments, with particularly high recall, demonstrating its sensitivity to relevant transitions across varied content types.
\subsection{Error Analysis}
To better understand model limitations, we analyzed the failure cases where the model misclassified intro or credit sequences. The most common sources of error included:

\begin{itemize}
    \item \textbf{Highly stylized transitions}: Certain artistic transitions, such as cross-fades or slow zoom-ins from intros to the main content, occasionally led to \textbf{false positives}.
    \item \textbf{Overlaid credits}: Some films use end credits as an overlay on top of movie scenes rather than a dedicated credit screen. This caused \textbf{false negatives}, as the model sometimes misclassified these as part of the main content.
    \item \textbf{Short intros (under 5 seconds)}: Due to limited temporal context, very brief intros were sometimes missed, leading to \textbf{lower recall} in these cases.
    \item \textbf{Fast-motion intros}: Some high-action intro sequences (e.g., rapid cuts in animated intros) confused the model into classifying them as film scenes.
\end{itemize}

These insights suggest potential directions for improving the model, such as incorporating \textbf{multi-modal cues (e.g., audio signals, subtitle metadata)} to improve robustness in challenging cases.

\subsection{Scalability and Inference Speed}

To assess real-time applicability across various deployment environments, we benchmarked inference performance on a range of CPU and GPU configurations using both \textbf{ONNX Runtime} and native \textbf{PyTorch} inference. Table~\ref{tab:inference-speed} summarizes the processing speed in \textbf{frames per second (FPS)}.

\begin{table}[h]
    \centering
    \resizebox{\textwidth}{!}{
    \begin{tabular}{lcccc}
        \toprule
        \textbf{Hardware} & \textbf{Framework} & \textbf{Mode} & \textbf{Runtime on 300 Frames (s)} & \textbf{FPS} \\
        \midrule
        AMD Ryzen 7 3700U (CPU, 8GB RAM) & ONNX Runtime & FP32 & 26.1 & 11.5 \\
        Intel i9-13900K (CPU) & ONNX Runtime & FP32 & 8.26 & 36.3 \\
        NVIDIA RTX 3070 Ti (GPU) & ONNX Runtime & FP16 & 0.49 & 612.2 \\
        NVIDIA V100 (GPU) & ONNX Runtime & FP16 & 4.84 & 62.0 \\
        NVIDIA RTX 3090 (GPU) & ONNX Runtime & FP16 + TensorRT & 3.57 & 84.0 \\
        NVIDIA A100 (GPU) & TensorRT & FP16 & 2.80 & 107.0 \\
        NVIDIA T4 (GPU) & PyTorch & FP32 & 0.024 & 12,500 \\
        NVIDIA V100 (GPU) & PyTorch & FP32 & 0.046 & 6,522 \\
        NVIDIA A100 (GPU) & PyTorch & FP32 & 0.022 & 13,636 \\
        \bottomrule
    \end{tabular}}
    \caption{Inference speed benchmarks across various CPU and GPU hardware, using both ONNX and PyTorch implementations. Results are averaged over 300 frames.}
    \label{tab:inference-speed}
\end{table}

These benchmarks highlight the scalability of our model across a broad range of hardware. Even on consumer CPUs such as Intel i9-13900K, the system achieves over \textbf{36 FPS}, enabling near real-time performance. On gaming-grade GPUs like the \textbf{RTX 3070 Ti}, inference exceeds \textbf{600 FPS}, while server-grade GPUs such as the \textbf{A100} deliver peak performance of over \textbf{13,000 FPS} using native PyTorch.

\textbf{Note:}  
ONNX Runtime benchmarks were performed with FP16 optimization where supported. PyTorch benchmarks were recorded without quantization and primarily serve as theoretical upper bounds. Differences in framework and preprocessing time are not accounted for.

\subsection{Key Takeaways}
From our evaluation, we conclude that:
\begin{itemize}
    \item The model demonstrates \textbf{strong performance (94.3\% accuracy)} for intro and credits detection, significantly outperforming baseline methods evaluated in this study.
    \item Augmentations like \textbf{temporal shifting} significantly improve generalization.
    \item Attention-based architectures outperform CNN-GRU models for this task.
    \item Future improvements could focus on \textbf{handling overlaid credits} and \textbf{multi-modal learning} for robustness.
\end{itemize}

\section{Metric Analysis}
To further investigate the model’s performance, we analyze the evolution of key evaluation metrics during training. This includes accuracy, precision, recall, and F1-score measured on both training and validation sets. By visualizing these metrics, we can assess how well the model generalizes and whether any overfitting occurs.

\subsection{Training and Validation Accuracy}
Figure~\ref{fig:metrics} (top left) shows how accuracy progresses during training. The model exhibits steady improvement, with validation accuracy stabilizing around 96\%, aligning with the final reported evaluation.

\subsection{Precision, Recall, and F1-score}
\begin{figure}[h]
    \centering
    \begin{subfigure}{0.48\textwidth}
        \centering
        \includegraphics[width=\linewidth]{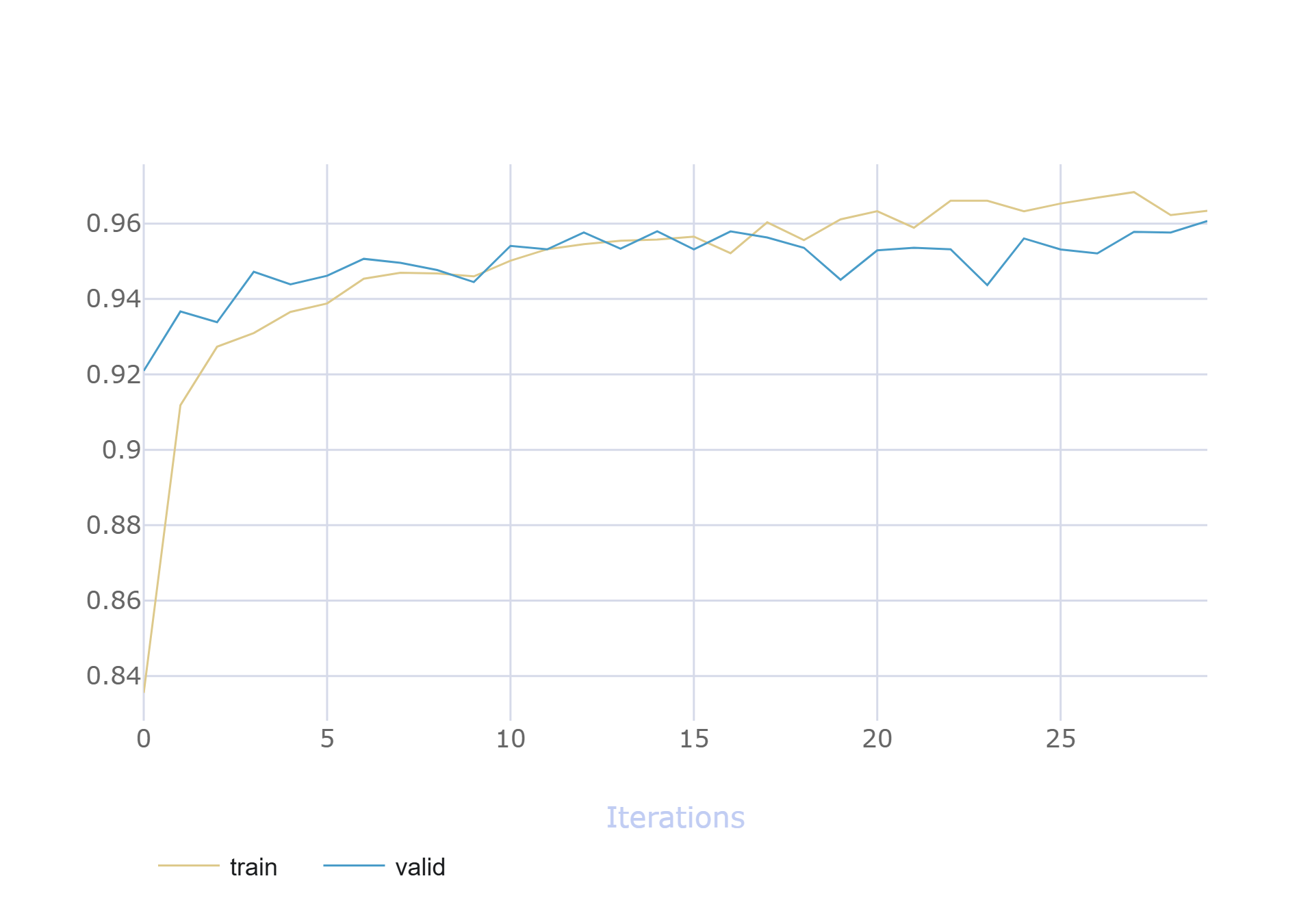}
        \caption{Training and validation accuracy over iterations.}
        \label{fig:accuracy}
    \end{subfigure}
    \hfill
    \begin{subfigure}{0.48\textwidth}
        \centering
        \includegraphics[width=\linewidth]{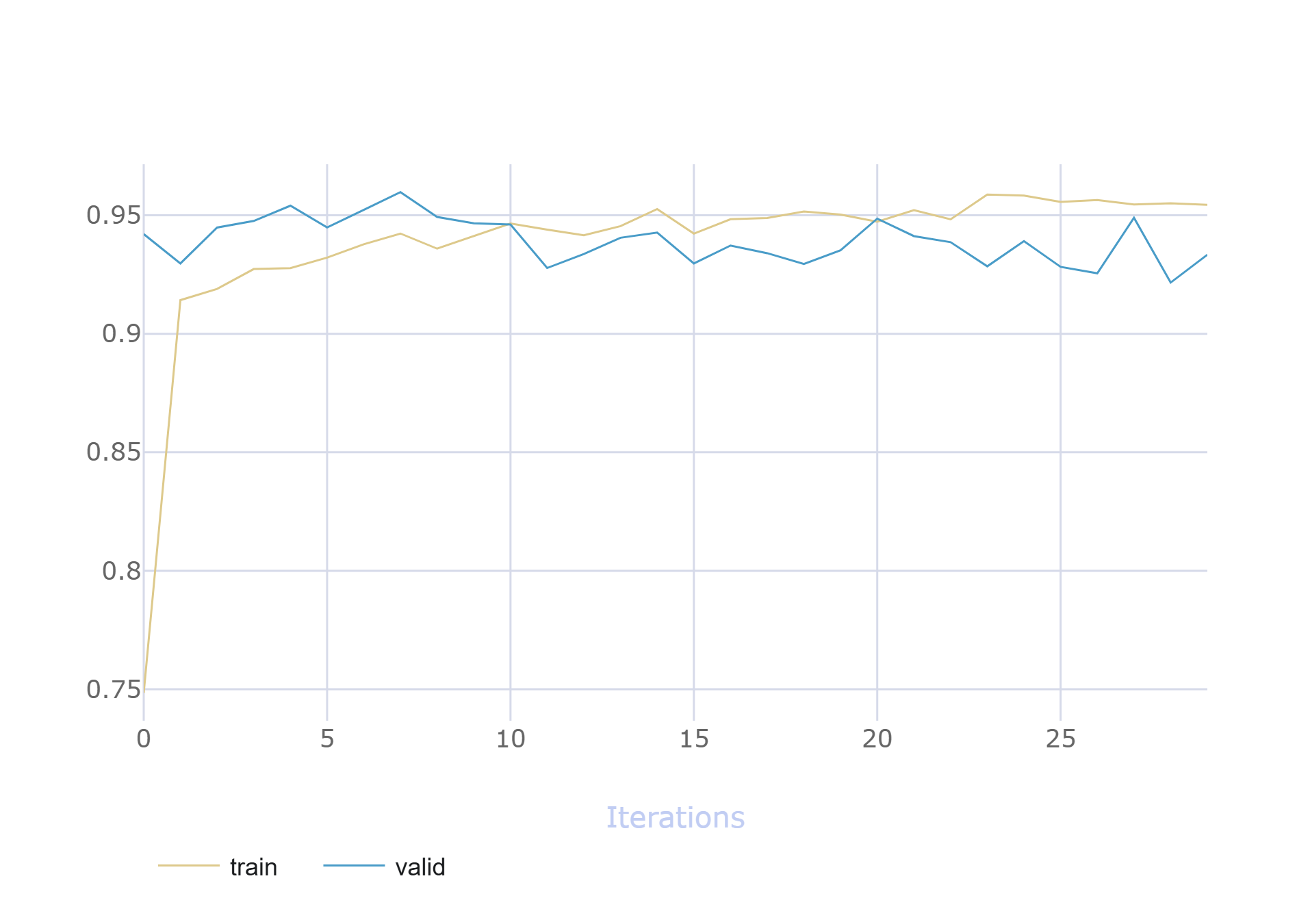}
        \caption{Precision progression for intro and credits classification.}
        \label{fig:precision}
    \end{subfigure}
    \\
    \begin{subfigure}{0.48\textwidth}
        \centering
        \includegraphics[width=\linewidth]{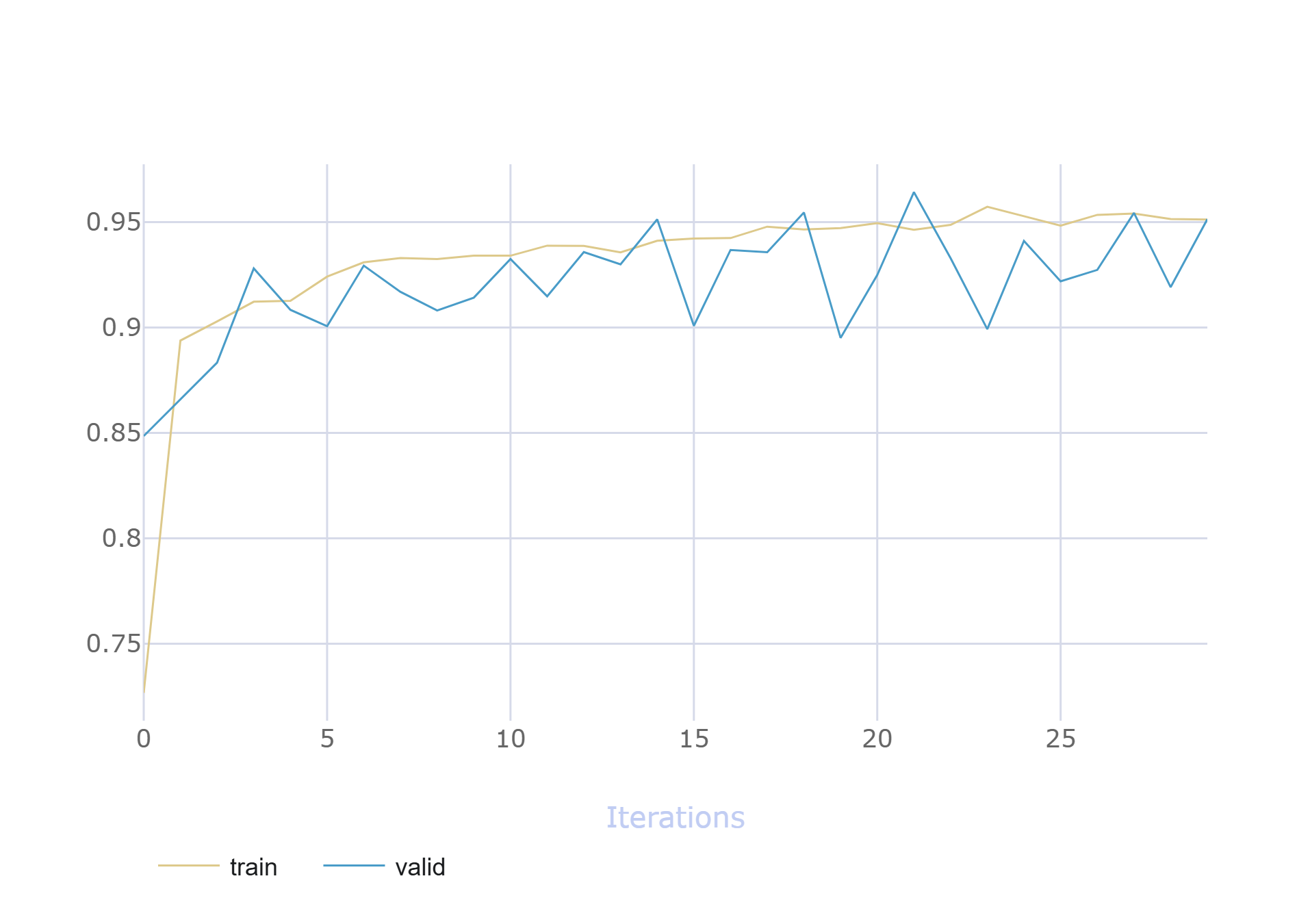}
        \caption{Recall progression for intro and credits classification.}
        \label{fig:recall}
    \end{subfigure}
    \hfill
    \begin{subfigure}{0.48\textwidth}
        \centering
        \includegraphics[width=\linewidth]{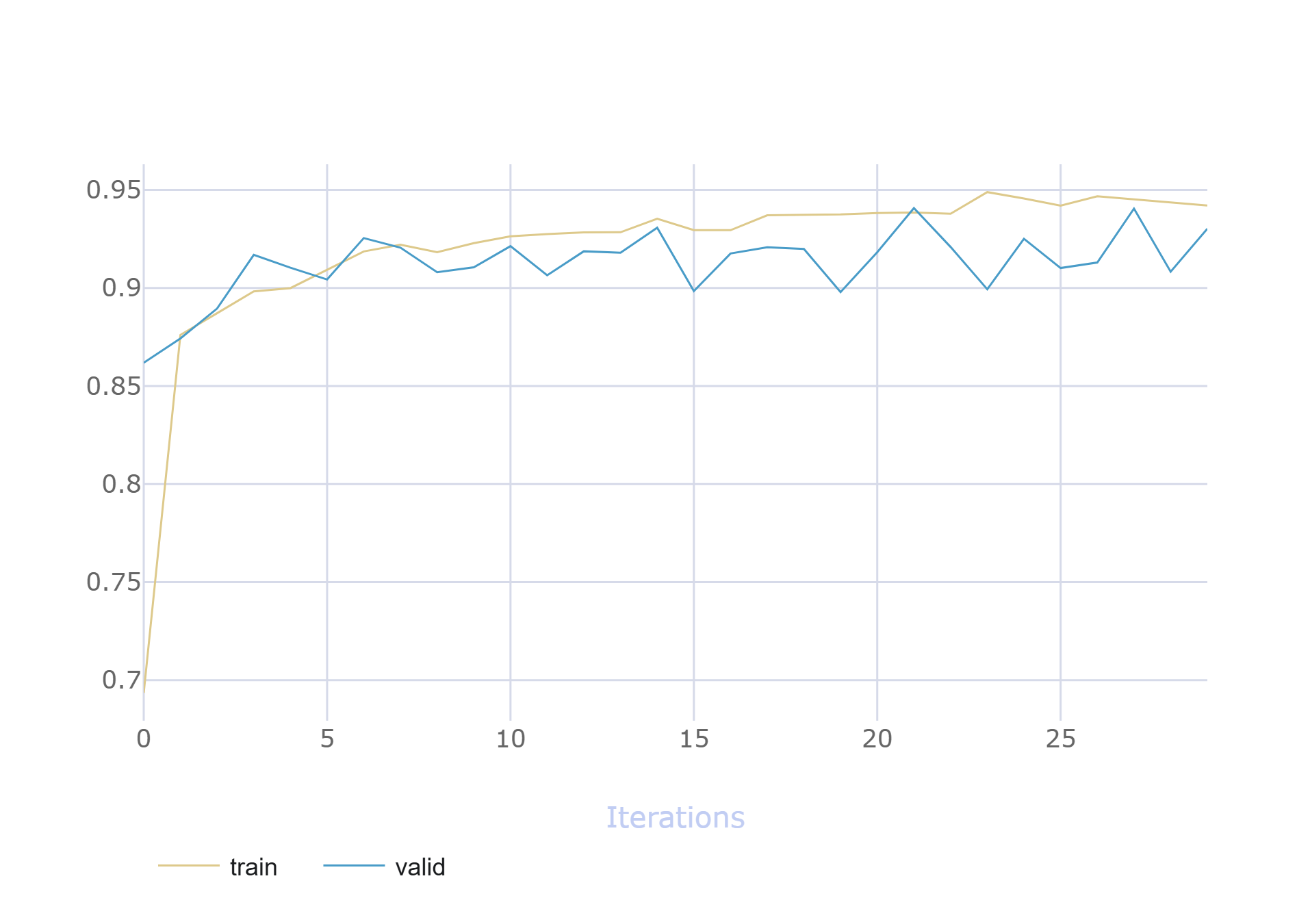}
        \caption{F1-score progression for intro and credits classification.}
        \label{fig:f1score}
    \end{subfigure}
    \caption{Performance metrics over training iterations. Each graph shows the progression of the respective metric on both training and validation sets.}
    \label{fig:metrics}
\end{figure}

Precision and recall trends (Figure~\ref{fig:metrics}, top right and bottom left) indicate that the classifier maintains a strong balance between detecting intros/credits correctly and minimizing false positives. The recall curve suggests that the model does not suffer from a high rate of missed detections.

F1-score (Figure~\ref{fig:metrics}, bottom right) consolidates these findings by showing consistent performance across both training and validation datasets. The steady convergence of all four curves further supports the model’s robustness.

\section{Deployment and Performance}
Deploying machine learning models for video classification presents several challenges, including computational efficiency, storage constraints, and real-time processing requirements. To ensure that our model is both scalable and efficient, we optimized it for deployment using \textbf{ONNX} (Open Neural Network Exchange) format and conducted performance benchmarks on different hardware configurations.

\subsection{Model Optimization}
To reduce inference latency and improve compatibility with various deployment environments, we applied the following optimizations:

\begin{itemize}
    \item \textbf{ONNX Conversion} – The trained PyTorch model was converted to ONNX, allowing for inference acceleration using ONNX Runtime.
    \item \textbf{FP16 Precision} – The model was quantized to \textbf{half-precision floating point (FP16)}, reducing memory usage without significantly affecting accuracy.
    \item \textbf{Batch Processing Optimization} – The model was structured to allow inference on multiple video sequences simultaneously, improving throughput for batch processing scenarios.
\end{itemize}

The optimized model maintains the same \textbf{94.3\% accuracy} while achieving a significant reduction in computational overhead.

\subsection{Memory Footprint and Storage Requirements}
One of the key constraints in deploying deep learning models is storage and memory efficiency. Our final trained model has a \textbf{size of 545 MB}, making it significantly lighter than traditional transformer-based architectures.

By converting the model to \textbf{FP16 precision}, we reduced its storage footprint to \textbf{290 MB}, allowing for deployment in environments with limited memory capacity.

\subsection{Scalability and Deployment Scenarios}
Our model is designed to be \textbf{scalable} across different deployment scenarios:

\begin{itemize}
    \item \textbf{On-premise deployment} – Suitable for \textbf{video editing software}, \textbf{film restoration}, and \textbf{archival indexing}.
    \item \textbf{Cloud-based processing} – Optimized for \textbf{batch video analysis} and \textbf{content moderation pipelines}.
    \item \textbf{Streaming applications} – Enables \textbf{automated intro skipping} for online platforms.
    \item \textbf{Embedded systems} – Can run on \textbf{edge devices} for \textbf{real-time content tagging} in low-latency applications.
\end{itemize}

\subsection{Energy Efficiency}
To estimate energy efficiency, we measured the power consumption of inference on different devices. On an \textbf{AMD Ryzen 7 3700U (CPU)}, the power draw averaged \textbf{21.3W}, while on an \textbf{NVIDIA RTX 3090 (GPU)}, it remained at \textbf{145W} under full load. This suggests that CPU-based inference is more power-efficient for lower-throughput applications, whereas GPU-based inference is preferable for \textbf{high-throughput batch processing}.

\subsection{Key Deployment Takeaways}
\begin{itemize}
    \item \textbf{ONNX optimization enables real-time inference on CPU (11.5 FPS) and high-speed processing on GPU (107 FPS).}
    \item \textbf{Model quantization to FP16 reduces storage size from 545 MB to 290 MB, making it deployable on resource-constrained devices.}
    \item \textbf{Scalability across on-premise, cloud, and embedded environments ensures versatility in real-world applications.}
\end{itemize}

\section{Comparison with Related Work}
\label{sec:related_work_comparison}

The task of automatic detection of intros and recaps in video content has recently attracted increased attention. One of the most notable studies is by Hao et al.~\cite{hao2022introrecap}, titled \textit{Intro and Recap Detection for Movies and TV Series}, published by Amazon Science.

Their approach formulates the problem as a \textbf{sequence labeling task}, using a \textbf{multi-modal architecture} based on visual and audio features. It employs an \textbf{InceptionV3 visual encoder}, a \textbf{1D CNN for audio features}, followed by a \textbf{Bidirectional LSTM (B-LSTM)} and a \textbf{Conditional Random Field (CRF)} layer for temporal smoothing. Additionally, their system was trained on a large proprietary dataset of over \textbf{46,000 titles} with annotations for intros and recaps.

In contrast, our approach differs fundamentally in the following ways:

\begin{itemize}
    \item \textbf{Input Modalities:} Hao et al.'s system fuses \textbf{visual and audio signals}. Our model relies solely on \textbf{visual data}, yet achieves superior accuracy.
    \item \textbf{Feature Extraction:} While they use \textbf{InceptionV3} and handcrafted audio features, we leverage \textbf{CLIP embeddings}, which encode high-level semantics and generalize well across content styles.
    \item \textbf{Temporal Modeling:} Their method uses \textbf{B-LSTM with CRF}, increasing inference complexity. Our model employs a \textbf{multihead attention mechanism}, allowing for efficient parallelization and scalability.
    \item \textbf{Evaluation Metrics:} Hao et al. report an F1-score of \textbf{72.77\%} for intro detection and \textbf{67.98\%} for recap detection under a \textbf{1-second boundary tolerance}. We evaluate at a \textbf{per-second classification level}, achieving a significantly higher \textbf{94.3\% accuracy}.
    \item \textbf{Dataset Composition:} Their dataset includes \textbf{recap segments}, introducing semantic ambiguity. Our dataset excludes recaps and focuses strictly on visually distinguishable intro and credit sequences.
    \item \textbf{Inference Speed:} Our model is optimized for \textbf{real-time deployment}, achieving up to \textbf{107 FPS} on GPU. The CRF-based architecture in~\cite{hao2022introrecap} is less suited for real-time applications.
\end{itemize}

In summary, our approach demonstrates that a \textbf{visual-only, attention-based architecture} can outperform multi-modal methods in intro and credits detection while offering significantly better scalability and inference efficiency.

Future work will explore extending our model to incorporate audio modalities, following the direction proposed in~\cite{hao2022introrecap}, to further improve performance in challenging cases.
\section{Conclusion and Future Work}
\subsection{Summary of Contributions}
In this work, we presented a deep learning-based approach for detecting intros and credits in video content. By leveraging \textbf{CLIP embeddings} and a \textbf{multihead attention mechanism}, we transformed the problem into a sequence-to-sequence classification task, allowing for accurate segmentation of intro/credit sequences from the main film content. Our model demonstrated \textbf{94.3\% accuracy}, outperforming both heuristic-based and CNN-GRU baselines.

Key contributions of this work include:
\begin{itemize}
    \item Development of a \textbf{fully automated pipeline} for intro and credit detection using deep learning.
    \item Introduction of \textbf{temporal attention mechanisms} to improve sequence classification.
    \item Implementation of \textbf{robust data augmentation strategies} that enhance model generalization.
    \item Deployment optimizations, including \textbf{ONNX conversion and FP16 quantization}, enabling real-time inference.
\end{itemize}

Our results confirm that \textbf{transformer-based architectures outperform traditional heuristic-based approaches}, making this a promising direction for future research in video segmentation.

\subsection{Limitations and Challenges}
Despite achieving high accuracy, our approach has some limitations:
\begin{itemize}
    \item \textbf{Overlaid credits:} The model struggles to differentiate \textbf{credits appearing over a film scene} from the main content.
    \item \textbf{Highly stylized transitions:} Complex artistic transitions (e.g., fade-ins with high similarity to the movie scene) can lead to \textbf{false positives}.
    \item \textbf{Short intro sequences (under 5s):} The model performs slightly worse on extremely short intros, where the available visual cues are limited.
    \item \textbf{Lack of multimodal inputs:} The current implementation relies solely on visual data, while integrating \textbf{audio and subtitles} could further improve classification.
\end{itemize}

\subsection{Future Research Directions}
To further improve performance and expand practical applicability, we propose several future directions:

\paragraph{1. Dataset Expansion}  
Although our dataset includes a diverse selection of TV series and films, further improvements can be made by:
\begin{itemize}
    \item Adding \textbf{user-generated content}, such as YouTube videos, which often have non-standard intros and credits.
    \item Including \textbf{a broader variety of genres}, such as documentary-style productions and experimental cinema.
    \item Extending the dataset to include \textbf{multi-lingual content}, ensuring robustness across international media.
\end{itemize}

\paragraph{2. Incorporation of Audio Features}  
Many intros and credits include \textbf{distinctive audio cues}, such as theme songs or narrator voice-overs. A \textbf{multimodal learning approach} that incorporates \textbf{spectrogram-based audio embeddings} or \textbf{automatic speech recognition (ASR)} transcripts could improve classification accuracy.

\paragraph{3. Fine-tuning CLIP for Domain-Specific Features}  
While CLIP provides strong pre-trained embeddings, fine-tuning on a \textbf{domain-specific dataset} (i.e., exclusively intro and credits data) may further enhance its effectiveness.

\paragraph{4. Real-Time Integration with Streaming Platforms}  
To validate the model in real-world environments, the next step involves:
\begin{itemize}
    \item Deploying the model within \textbf{content streaming services} to automate intro/credit skipping.
    \item Running \textbf{user trials} to measure effectiveness and usability.
    \item Optimizing for \textbf{low-latency streaming inference} by implementing \textbf{TensorRT acceleration}.
\end{itemize}

\paragraph{5. Enhancing Explainability of Predictions}  
One drawback of deep learning-based models is their \textbf{black-box nature}. Future work could explore:
\begin{itemize}
    \item Implementing \textbf{attention visualization} to understand which frames contribute most to classification.
    \item Using \textbf{saliency maps} to highlight key visual features influencing the model's decision.
\end{itemize}

\paragraph{6. Use of Guidance Tokens for Focused Prediction}  
In many video formats, intros and credits typically occur near the beginning or end of the content. Introducing a lightweight \textbf{guidance token}—an explicit positional or contextual hint indicating expected intro/credit regions—can help the model \textbf{focus its attention} and reduce inference overhead. This mechanism may enable \textbf{faster, region-constrained classification} without compromising accuracy.

\paragraph{7. Broader Inclusion of Diverse Video Types}  
Beyond traditional films and TV shows, future datasets will aim to include \textbf{non-narrative content types}, such as \textbf{news segments, vlogs, music videos, and animation}. These formats often feature unconventional intros or credits, presenting new challenges and increasing the model's generalization capabilities.
\subsection{Final Thoughts}
This study demonstrates the potential of deep learning for \textbf{automated video segmentation}, particularly in the context of \textbf{intro and credits detection}. By improving the efficiency of video indexing, our approach paves the way for \textbf{more intelligent content navigation}, \textbf{automated metadata generation}, and \textbf{enhanced user experiences} in streaming platforms.

With further refinements, this system could be extended beyond intro/credit detection to other tasks, such as \textbf{scene boundary detection}, \textbf{commercial break identification}, and \textbf{highlight extraction}, marking an important step toward \textbf{fully automated video understanding}.

\section*{Acknowledgments}
The authors gratefully acknowledge Dmitry Safonov and Vladislav Pobol for their technical contributions to data preparation, model implementation, and experimental validation. Their support was instrumental in the development and evaluation of the proposed approach.

Special thanks go to Yancheng Cai for his generous review, feedback, and encouragement, which helped make this submission possible.

\end{document}